\documentclass[runningheads]{llncs}
\usepackage{graphicx}
\usepackage{amsmath}
\usepackage{amsfonts}
\usepackage{xcolor}
\usepackage{subcaption}
\usepackage{url}

\usepackage{marginnote}

\newcommand\defeq{\mathrel{\stackrel{\makebox[0pt]{\mbox{\normalfont\scriptsize def}}}{:=}}}

\newcommand{\R}{\mathbb{R}}

\newcommand{\N}{\mathbb{N}}

\usepackage{amsmath} 
\usepackage{tikz}
\tikzset{>=latex} 
\usepackage{pgfplots} 
\usepackage{xcolor}
\usepackage[outline]{contour} 
\contourlength{1.2pt}
\usetikzlibrary{positioning,calc}
\usetikzlibrary{backgrounds}

\pgfmathdeclarefunction{gauss}{3}{%
  \pgfmathparse{1/(#3*sqrt(2*pi))*exp(-((#1-#2)^2)/(2*#3^2))}%
}
\pgfmathdeclarefunction{cdf}{3}{%
  \pgfmathparse{1/(1+exp(-0.07056*((#1-#2)/#3)^3 - 1.5976*(#1-#2)/#3))}%
}
\pgfmathdeclarefunction{fq}{3}{%
  \pgfmathparse{1/(sqrt(2*pi*#1))*exp(-(sqrt(#1)-#2/#3)^2/2)}%
}
\pgfmathdeclarefunction{fq0}{1}{%
  \pgfmathparse{1/(sqrt(2*pi*#1))*exp(-#1/2))}%
}

\colorlet{mydarkblue}{blue!30!black}

\usepgfplotslibrary{fillbetween}
\usetikzlibrary{patterns}
\pgfplotsset{compat=1.12} 

\def\N{50}

\begin{document}
\title{Runtime Monitoring DNN-based Perception}
\subtitle{(via the Lens of Formal Methods)}

\author{Chih-Hong Cheng\inst{1} \and
Michael Luttenberger\inst{1} \and
Rongjie Yan\inst{2}} 
\authorrunning{C.-H.~Cheng et al.}
%
\institute{Technical University of Munich, Germany \\
\email{chih-hong.cheng@tum.de, luttenbe@in.tum.de}\\ \and
Institute of Software, Chinese Academy of Sciences, China \\
\email{yrj@ios.ac.cn}}
\maketitle              
\begin{abstract}
Deep neural networks (DNNs) are instrumental in realizing complex perception systems. As many of these applications are safety-critical by design, engineering rigor is required to ensure that the functional insufficiency of the DNN-based perception is not the source of harm. In addition to conventional static verification and testing techniques employed during the design phase, there is a need for runtime verification techniques that can detect critical events, diagnose issues, and even enforce requirements. This tutorial aims to provide readers with a glimpse of techniques proposed in the literature. We start with classical methods proposed in the machine learning community, then highlight a few techniques proposed by the formal methods community. While we surely can observe similarities in the design of monitors, how the decision boundaries are created vary between the two communities. We conclude by highlighting the need to rigorously design monitors, where data availability outside the operational domain plays an important role.

\end{abstract}

\section{Introduction}
Deep neural networks (DNNs) play a pivotal role in realizing perception, revolutionizing our understanding of how machines can interpret and interact with the world. DNNs have transformed various fields, including computer vision, natural language processing, and audio recognition, by learning and extracting meaningful patterns from vast amounts of data. 
Apart from applications such as urban autonomous driving, it is now gaining attention in other domains such as railways and avionics. For example, the safe.trAIn project\footnote{\url{https://safetrain-projekt.de/en/}} considered how DNN-based perception can be used in creating driverless autonomous trains for regional transportation. Given the input represented in digital formats (RGB images or lidar point clouds), one can easily apply DNN to perform various tasks such as classification, 2D or 3D object detection, and semantic segmentation, to name a few applications.

Nevertheless, the use of DNNs in safety-critical applications such as autonomous driving necessitates additional care and precautionary measures, given the potential risks associated with incorrect or unreliable decisions made by these networks. In such applications, where human lives or important infrastructure are at stake, the robustness and reliability of deep neural networks become paramount.
One prominent technique to address these concerns is \emph{runtime monitoring}, which involves continuously monitoring the behavior and outputs of the network during its operation. Alongside runtime monitoring, other techniques like formal verification and thorough testing are also employed to ensure the safety and reliability of deep neural networks in critical domains. Combining these techniques enhances the trustworthiness and dependability of deep neural networks in safety-critical applications, providing a vital layer of assurance and mitigating potential risks.

\begin{figure}[t]
    \centering
    \includegraphics[width=\textwidth]{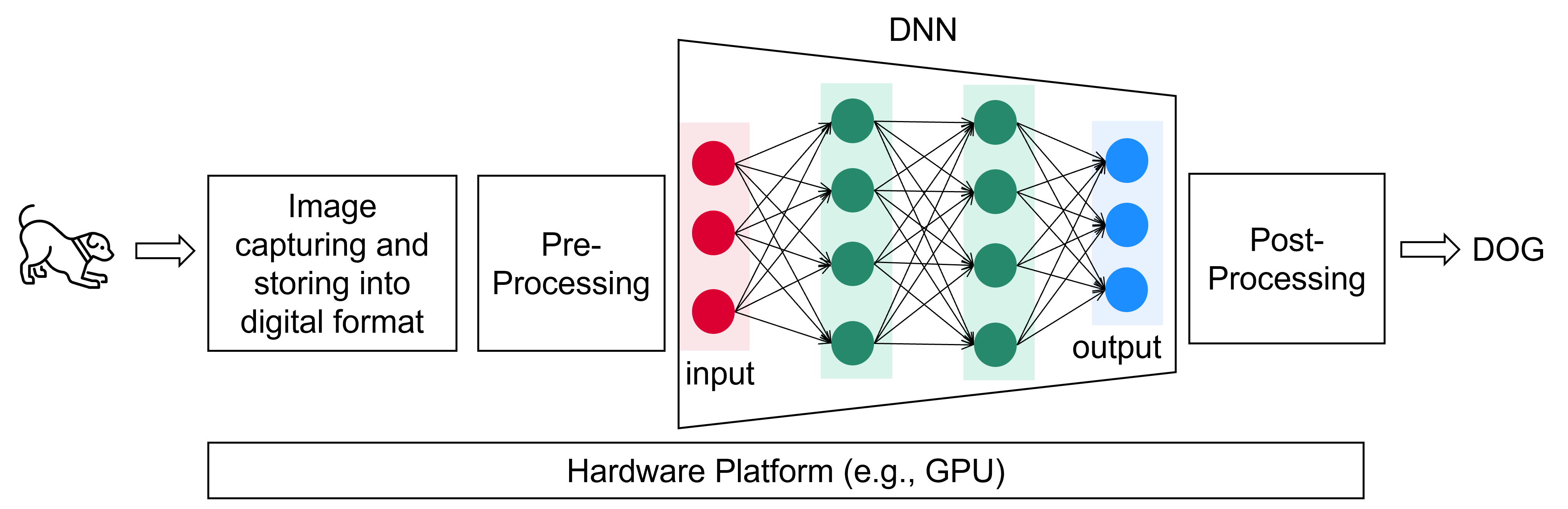}
    \caption{A typical DNN-based perception system}
    \label{fig:pipeline}
\end{figure}

This tutorial\footnote{All supplementary materials for the tutorial are made available at \url{https://sites.google.com/site/chengchihhong/home/teaching/rv23}} aims to offer readers a glimpse into the complex topic of runtime monitoring DNN-based perception systems, where active research has been conducted in the machine learning community and has been receiving attention in the formal methods community. As the topic is very broad, we would like to constrain the scope by considering the simplified perception system pipeline illustrated in Fig.~\ref{fig:pipeline}. The perception system first performs image formation and stores the image in a digital format. Then pre-processing methods can be applied before feeding into neural networks. Commonly used pre-processing methods include denoising or quantization. Post-processing refers to algorithms such as polishing the result of the DNN. For example, in object detection, \emph{non-max suppression} is used to remove spatially overlapping objects whose output probability is lower. The tutorial emphasizes the runtime monitoring of DNN and may partly include the pre-and post-processing. Regarding the monitoring of hardware platforms, although there exists some confirmation on the hardware faults influencing the result of prediction~\cite{qutub2022hardware}, in this tutorial, we have decided to neglect them and relegate the monitoring activities to classical safety engineering paradigms such as ISO 26262 for dealing with hardware faults.

\section{Challenges in Monitoring Perception Systems}

Monitoring perception systems poses some unique challenges in contrast to the standard runtime verification paradigm, which we detail in the following sections. 

\subsubsection{Specification} In runtime verification, the type of formal specifications of interest can be state properties (invariance), temporal properties characterizing trace behaviors utilizing temporal logics, like qualitative linear temporal logic (LTL)~\cite{pnueli1977temporal} or 
 extensions thereof like signal temporal logic (STL)~\cite{donze2010robust} or metric temporal logic (MTL)~\cite{koymans1990specifying}, and hyper-properties comparing behaviors of multiple traces. The definition of states is application-specific.

For learning-based perception systems such as object detection with input from images (pedestrian, car, bicycle), the undesired situation for the perception, when considering only the single input than a sequence of inputs,  can be separated into two parts:
\begin{itemize}
    \item The output can be incorrect in one of the following ways, namely
        \begin{enumerate}
        \item object not detected, 
        \item detecting ghost objects, 
        \item incorrect object classification (a pedestrian is detected as a car), or
        \item incorrect object size (too big or too small).
    \end{enumerate}

    \vspace{1mm}
    \item The input can also be ``strange" due to not resembling what is commonly expected as input (e.g., random noise rather than highway road scenes). 
\end{itemize}

In many systems, during the runtime verification process, one can access the state information and know if a state property is satisfied. The measured state information offers the ``ground truth". For perception, however, the \emph{ground truth (i.e., whether an object exists) is unavailable for comparison during run-time}. Therefore, the above-mentioned error types can not be directly detected. This ultimately leads to \emph{indirect} methods for error detection, with methods including the following types: 

\begin{itemize}
    \item error detection via \emph{redundancy} (e.g., comparing the prediction based on additional sensor modalities),
        \item error detection via \emph{domain knowledge} (e.g., utilizing Newton's law or classical computer vision methods to filter problematic situations),
    \item error detection via \emph{temporal consistency} (e.g., comparing the result of prediction over time), and
    \item error detection via \emph{monitoring the decision mechanism of the DNN}.
\end{itemize}

While the first three methods (redundancy, domain knowledge, temporal consistency) can also be applied in any non-learning-based algorithms (e.g., algorithms utilizing classical computer vision algorithms), the latter is unique for deep neural networks. 

Finally, we must admit that a formal specification is usually incomplete. Thus, only some important and necessary safety aspects might be formalized. For complex autonomous systems, the ``operational design domain (ODD)'', i.e., the input domain that the system is expected to operate, is often incompletely formalized and only implicitly given by means of the collected data set.

\subsubsection{Reaction time} The reaction time between the occurrence of a prediction error and the error being detected by the monitor is crucial. As a realistic example, consider a perception module for autonomous driving operated at 10~FPS.
Assume that the object is continuously not perceived starting at time $t$, and the perception error is only detected at time $t+\Delta$ where the car performs a full break. As the vehicle can at least travel during the time interval $[t, t+\Delta]$, it may create dangerous scenarios when $\Delta$ is too large where applying maximum break is insufficient to avoid a collision.

\section{Formulation}

Throughout the text, for a vector $\vec{v}$, we refer $\vec{v}_i$ as its $i$-th component. We create the simplest formulation of neural networks using \emph{multi-layer perceptron}. In our formulation, we also assume that the neural network has been trained, i.e., the parameters of the network are fixed. The concepts can also be applied in convolutional neural networks, residual networks, or transformers.  

\begin{figure}[t]
    \centering
    \includegraphics[width=0.45\textwidth]{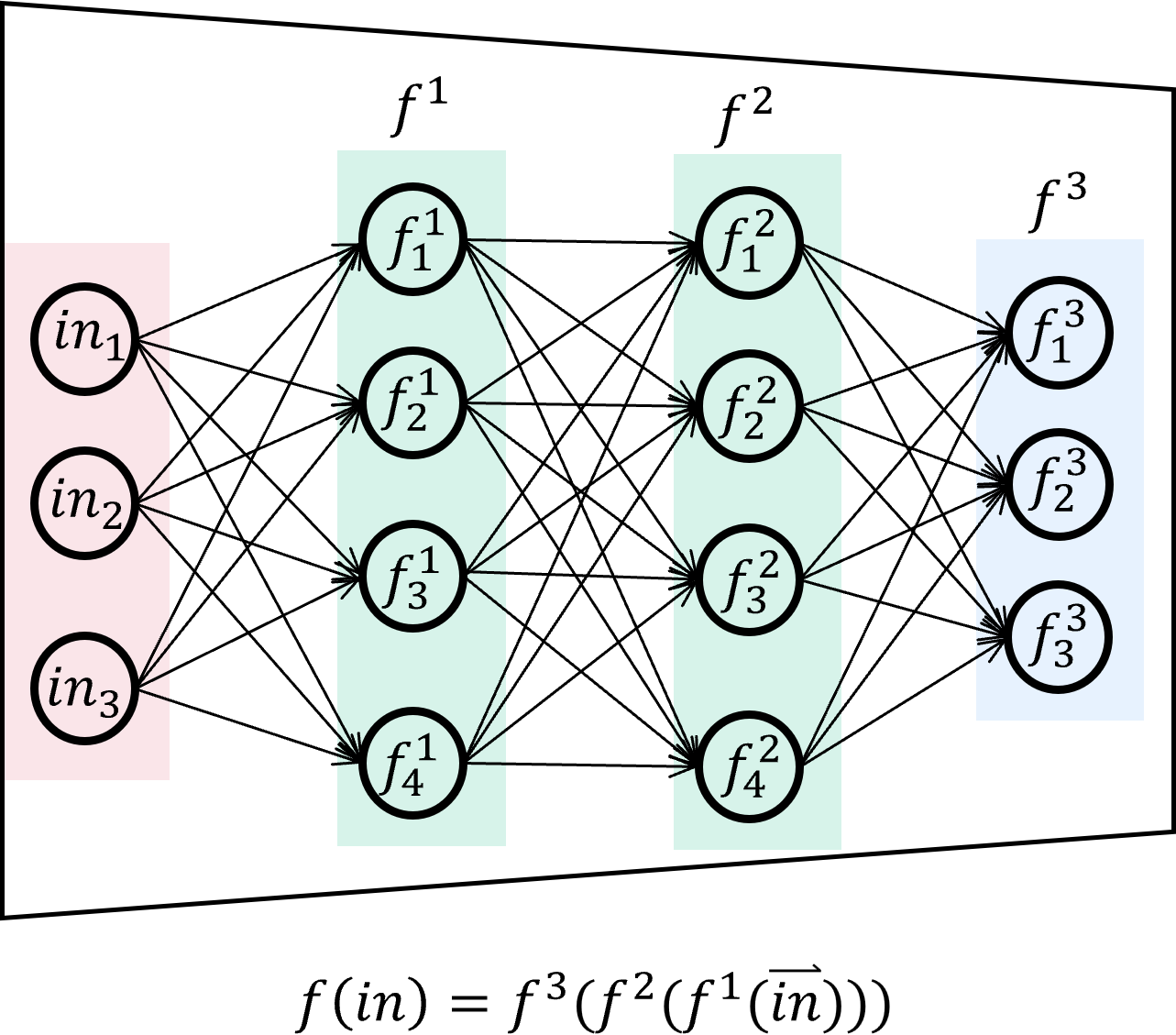}
    \caption{An example of multi-layer perceptron}
    \label{fig:neural.network.mlp}
\end{figure}

\begin{figure}[t]
    \centering
\includegraphics[width=0.9\textwidth]{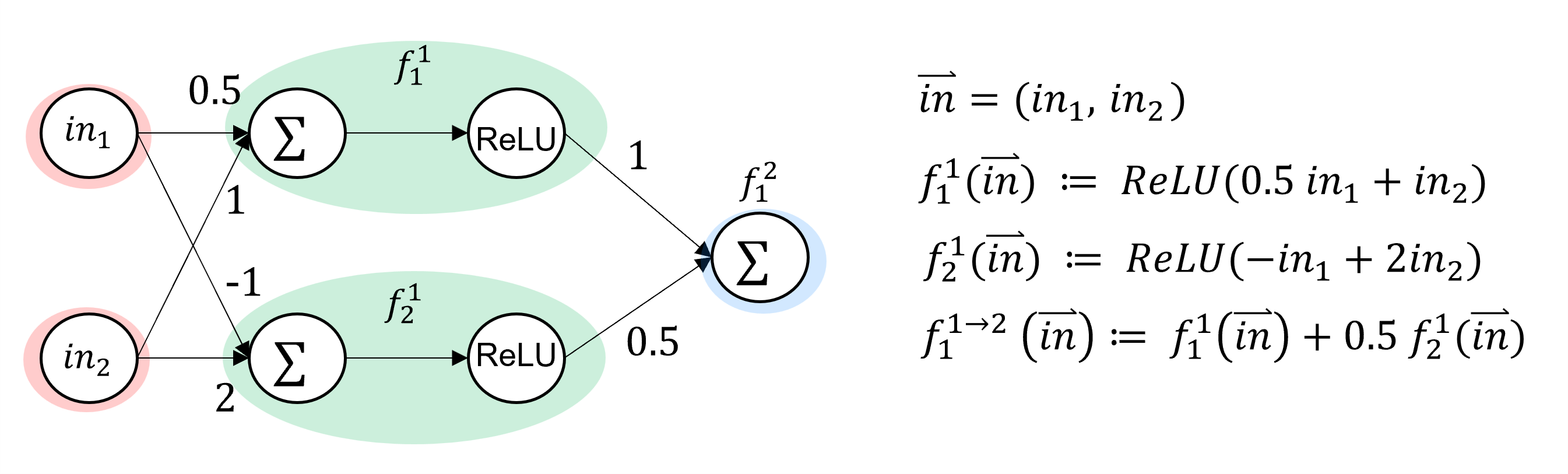}
    \caption{Forward propagation of a neural network}
    \label{fig:neural.network.forward.computation}
\end{figure}

A neural network $f$ is a composition of functions $f^L \circ f^{L-1} \circ \dots \circ f^2 \circ f^1$, where for $l \in \{1, \ldots, L\}$, $f^l: \R^{d_{l-1}} \rightarrow \R^{d_{l}}$ is the computational function of the $l$-th layer.  We refer to $d_0$ as the input dimension and $d_{L}$ as the output dimension. At layer $l$, there are $d_l$ neurons, with each neuron indexed $i$ being a function that takes the output from the previous layer and produces the $i$-th output of~$f^l$. In other words, $f^l_i: \R^{d_{l-1}} \rightarrow \R$. Given an input $\vec{in} \in \R^{d_0}$, the output of the neural network equals $f^{L}(f^{L-1}(\ldots (f^2(f^1(\vec{in})))))$. Finally, we use $f^{1\rightarrow l}(\vec{in})$ to abbreviate $f^{l}(f^{l-1}(\ldots (f^2(f^1(\vec{in})))))$, i.e., the computation of value with input being $\vec{in}$, and  use $f^{1\rightarrow l}_i(\vec{in})$ to extract the $i$-th neuron value. Fig.~\ref{fig:neural.network.mlp} illustrates an example of a neural network of~$3$ layers, taking a vector ($in_1$, $in_2$, $in_3$) 
 of~$3$ numerical values as input and producing~$3$ outputs. The computation of a neuron in a multi-layer perception is done by performing a weighted sum, followed by a non-linear activation function. Fig.~\ref{fig:neural.network.forward.computation} shows an example of how the computation is defined, where the \emph{ReLU activation function} (also simply \emph{ramp function}) is defined as follows:

\[
ReLU(x) \defeq \max(0,x) = H(x)\cdot x =
\begin{cases}
  x & \text{if $x> 0$} \\
  0 & \text{otherwise}
\end{cases}
\] 
with $H$ the \emph{Heaviside step function}. Fixed-point systems $x_1=f_1(x)\wedge \ldots\wedge x_n=f_n(x)$ ($x=(x_1,\ldots,x_n)\in\R^n, f_i\colon\R^n\to\R$) where $f_i$ is a composition of linear forms, $\max$ and $\min$ have been studied also as abstractions of dynamic systems before, e.g.\ as abstractions of programs or turn-based two-player games (see e.g.~\cite{DBLP:journals/toplas/GawlitzaS11}).

Finally, let $\mathcal{D}\defeq \{(\vec{in}, \vec{label})\}$ be the data set collected from the human-specified operational domain for training the neural network~$f$, where $\vec{in} \in \mathbb{R}^{d_0}$ and $\vec{label} \in \mathbb{R}^{d_L}$. 
 Let $\mathcal{D}^{train}, \mathcal{D}^{val}, \mathcal{D}^{test} \subseteq \mathcal{D}$ be the training, validation, and test set. 
   Commonly, $\mathcal{D}^{train} \cap \mathcal{D}^{val} = \mathcal{D}^{train} \cap \mathcal{D}^{test} = \mathcal{D}^{test} \cap \mathcal{D}^{val} = \emptyset$.

\section{Techniques}

\subsection{DNN monitoring techniques from the ML community}\label{subsec.monitoring.ml}

\begin{figure}[t]
    \centering
\includegraphics[width=0.7\textwidth]{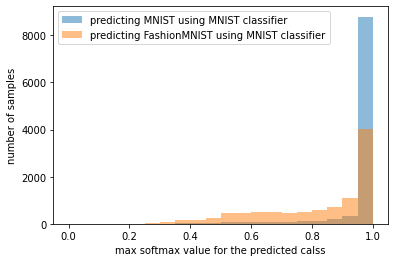}
    \caption{Using an MNIST classifier to predict the output from images in the MNIST and Fashion-MNIST data set. }
    \label{fig:neural.network.overly.confident}
\end{figure}

This section gives insight into some renowned approaches proposed by the ML community, intending to identify inputs that are \textbf{\emph{out-of-distribution}} (OoD), i.e., input data that are not similar to the data used in training\footnote{This concept is ambiguously defined in the field of machine learning.}. We start with the work of Hendrycks and Gimpel that introduces extra components (with a reconstruction component) and softmax decision~\cite{hendrycks2016baseline}, followed by the work of Liang, Li, and Srikant that uses temperature scaling and positively perturbed examples~\cite{liang2018enhancing}, and conclude the section with the work from Lee et~al. utilizing Mahalanobis distance~\cite{lee2018simple}.

\subsubsection{Limitations of the softmax output} 

In the case of classification, the final layer consists of the application of the so-called \emph{softmax function} which use the isomorphism from $\R$ to $\R_{>0}$ given by the exponentiation followed by the standard mean to normalize the output to a distribution over the possible classes. Formally, the $i$-th output of applying softmax is defined as follows.

\[
\sigma(\vec{x})_i \defeq \frac{e^{\vec{x}_i}}{\sum^{d_L}_{i=1} e^{\vec{x}_i}} 
\]

In probability theory, the output of the softmax function can be used to represent a categorical distribution, i.e., a probability distribution over~$d_L$ different possible outcomes.

Unfortunately, the work from Hendrycks and Gimpel~\cite{hendrycks2016baseline} discovered that pre-trained neural networks could be overconfident to out-of-distribution examples, thereby limiting the usefulness of directly using softmax as the output. To understand this concept, we have trained multiple neural networks on the MNIST data set~\cite{lecun1998mnist} for digit recognition. Subsequently, we use the trained neural network to classify clothing types for the Fashion-MNIST data set~\cite{Xiao2017FMNIST}. Ideally, we hope that the output of the softmax function should not generate values close to~$1$. Unfortunately, as illustrated in Fig.~\ref{fig:neural.network.overly.confident}, one can observe that among 10000 images from the Fashion-MNIST data set, around 4000 examples are predicted by the MNIST-classifier with a probability between 0.95 and 1.

Consequently, the authors in~\cite{hendrycks2016baseline} trained an additional classifier to detect if an input is out-of-distribution via the architecture as illustrated in Fig.~\ref{fig:baseline}. 
\begin{itemize}
    \item First, train a standard classifier ($f \defeq f^3 \circ f^2 \circ f^1$) with in-distribution data.

\begin{figure}[t]
    \centering
\includegraphics[width=0.6\textwidth]{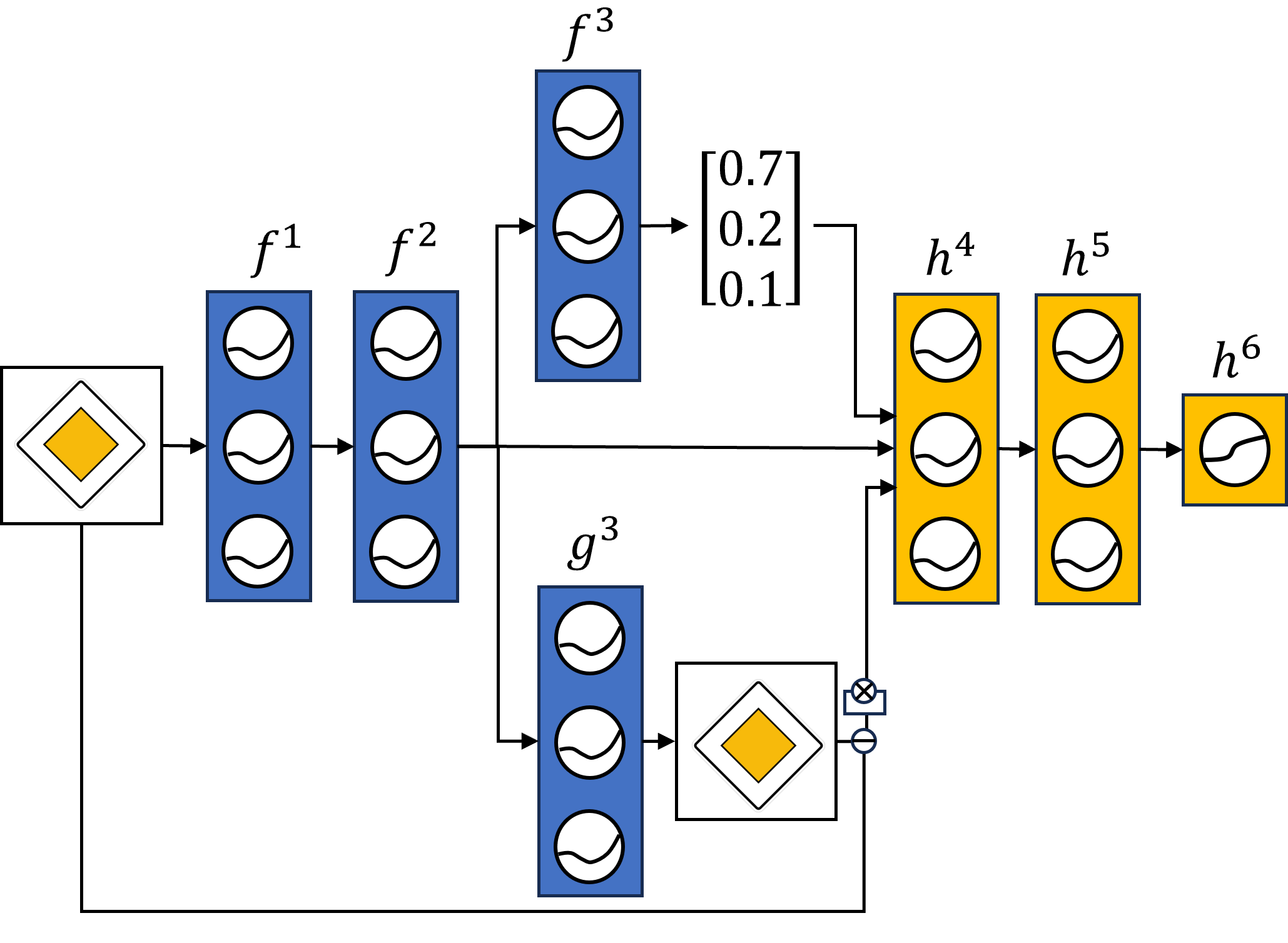}
    \caption{The architectural diagram for the OoD detector  by  Hendrycks and Gimpel~\cite{hendrycks2016baseline}. The blue layers ($f^1, f^2, f^3, g^3$) are trained with in-distribution data, and the yellow layers ($h^4, f^5, h^6$) are trained with both in-distribution and out-of-distribution data.}
    \label{fig:baseline}
\end{figure}

\begin{figure}[t]
\begin{subfigure}{.5\textwidth}
  \centering
  \includegraphics[width=.95\linewidth]{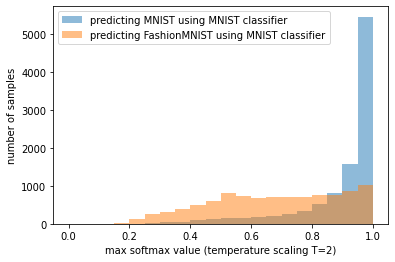}
  \caption{$T=2$}
  \label{fig:sfig1}
\end{subfigure}%
\begin{subfigure}{.5\textwidth}
  \centering
  \includegraphics[width=.95\linewidth]{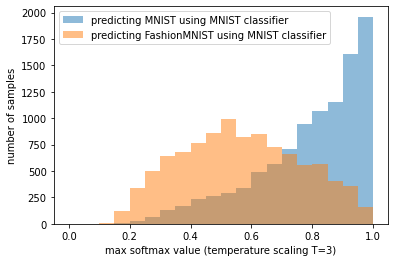}
  \caption{$T=3$}
  \label{fig:sfig2}
\end{subfigure}
\begin{subfigure}{.5\textwidth}
  \centering
  \includegraphics[width=.95\linewidth]{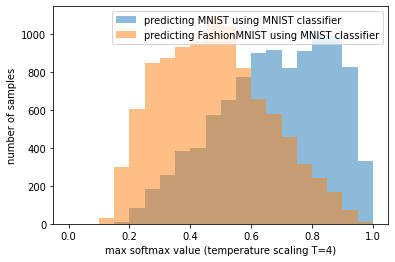}
  \caption{$T=4$}
  \label{fig:sfig3}
\end{subfigure}%
\begin{subfigure}{.5\textwidth}
  \centering
  \includegraphics[width=.95\linewidth]{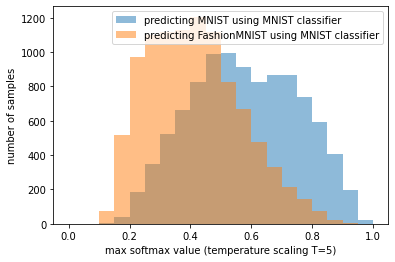}
  \caption{$T=5$}
  \label{fig:sfig4}
\end{subfigure}
    \caption{Applying temperature scaling on the same predictor for creating Fig.~\ref{fig:neural.network.overly.confident}}
\label{fig:neural.network.temperature.scaling}
\end{figure}

    \item Next, use the feature extractors of the predictor ($f^1, f^2$) to train another image reconstructor $g^3$, also with in-distribution data.

\item Finally, train the OoD detector $h^6 \circ h^5 \circ h^4$ with both in-distribution and out-of-distribution data, based on three types of inputs, namely
\begin{itemize}
    \item output of features~$f^2$,
    \item the softmax prediction from the standard classifier~$f^3$, and
    \item the quantity of the reconstruction error.  
\end{itemize}

\end{itemize}

Intuitively, if an input is out-of-distribution, it may have different features, or it may not be reconstructed, thereby hinting at the usefulness of these types of inputs for~$h^4$ in Fig.~\ref{fig:baseline}.

\subsubsection{Temperature scaling and gradient ascent} While the first method is based on the observation where softmax output can not be directly used as a means to detect out-of-distribution inputs, the immediate question turns to be if there exist lightweight methods to avoid training additional components. The ODIN approach by Liang, Li, Srikant~\cite{liang2018enhancing}  used temperature scaling and gradient ascent on inputs in-distribution to create better \emph{separators} between in-distribution and out-of-distribution input. Temperature scaling, as defined in Eq.~\ref{eq.temperature.scaling}, uniformly divide each dimension of $\vec{x}$ with a positive constant $T$ whose value is larger than~$1$, before performing the standard softmax function. Intuitively, it smooths out the large values caused by applying an exponential function. Fig.~\ref{fig:neural.network.temperature.scaling} shows the result of applying the temperature scaling (with different $T$ values) on the same predictor used in Fig.~\ref{fig:neural.network.overly.confident}. When $T=3$, when we set the OoD cutoff threshold to be~$0.6$ (i.e., if the largest output value is below~$0.6$, report OoD; otherwise, return the class that has the largest output value), we can filter a great part of out-of-distribution samples while maintaining the performance for in-distribution samples. 

\begin{equation}\label{eq.temperature.scaling}
    \sigma(\vec{x}, T)_i \defeq \frac{e^{\vec{x}_i / T}}{\sum^{d_L}_{i=1} e^{\vec{x}_i / T}} 
\end{equation}

The second step of the ODIN approach is to create a modified input $\vec{in}'$ which perturbs the input $\vec{in}$ to increase the prediction score of the largest class.  The intuition is that any input in-distribution is more likely to be perturbed to increase the maximum output probability. This technique needs to compute the gradient of the largest output over the input (i.e., $\frac{\partial \; f_i(\vec{in})}{\partial \;\vec{in}}$), and take a small step in the direction suggested by the gradient (thereby calling gradient ascent). This technique is the \emph{dual of adversarial perturbation}, which tries to decrease the prediction score of the largest class. 

\subsubsection{Mahalanobis distance}

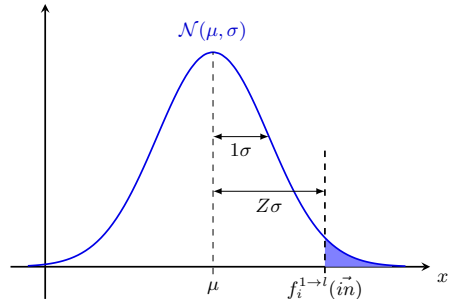
\begin{figure}[t]
    \centering

    \resizebox{.5\linewidth}{!}{
\begin{tikzpicture}[inner frame sep=0]
  \message{Normal distrubution p-value^^J}
  
  \def\q{5};
  \def\B{3};
  \def\S{8};
  \def\Bs{1.0};
  \def\Ss{1.5};
  \def\xmax{\B+3.2*\Bs};
  \def\ymin{{-0.15*gauss(\B,\B,\Bs)}};
  
  \begin{axis}[every axis plot post/.append style={
               mark=none,domain={-0.05*(\xmax)}:{1.08*\xmax},samples=\N,smooth},
               xmin={-0.1*(\xmax)}, xmax=\xmax,
               ymin=\ymin, ymax={1.1*gauss(\B,\B,\Bs)},
               axis lines=middle,
               axis line style=thick,
               enlargelimits=upper, 
               ticks=none,
               xlabel=$x$,
               every axis x label/.style={at={(current axis.right of origin)},anchor=north west},
               y=250pt
              ]
    
    \addplot[name path=B,thick,black!10!blue] {gauss(x,\B,\Bs)};
    \addplot[black,dashed,thick]
      coordinates {(\q,{-0.03*gauss(\B,\B,\Bs)}) (\q, {4.0*gauss(\q,\B,\Bs)})}
      node[below=-2pt,pos=0] {$f^{1\rightarrow l}_i(\vec{in})$};
    \addplot[black,dashed,thin]
      coordinates {(\B,{-0.035*gauss(\B,\B,\Bs)}) (\B, {gauss(\B,\B,\Bs)})}
      node[below=0pt,pos=0] {$\mu$};
    \addplot[<->,black,thin]
      coordinates {(\B,{gauss(\B-\Bs,\B,\Bs)}) (\B+\Bs, {gauss(\B+\Bs,\B,\Bs)})}
      node[below,midway] {$1\sigma$};
    \addplot[<->,black,thin]
      coordinates {(\B,{2.6*gauss(\q,\B,\Bs)}) (\q,{2.6*gauss(\q,\B,\Bs)})}
      node[below,midway] {$Z\sigma$};
    
    \path[name path=xaxis]
      (0,0) -- (\xmax,0);
    \addplot[white!50!blue] fill between[of=xaxis and B, soft clip={domain=\q:\xmax}];
    
    \node[above=2pt,  black!20!blue]    at (       \B,     {gauss(\B,\B,\Bs)}) {$\mathcal{N}(\mu,\sigma)$};
    
  \end{axis}
\end{tikzpicture}
}
    \caption{Using Z-score as a method for rejecting inputs with neuron values deviating largely from the mean}
    \label{fig:standard.deviation}
\end{figure}

\begin{figure}[t]
    \centering
\includegraphics[width=0.35\textwidth]{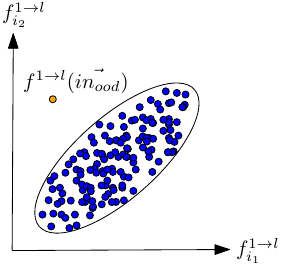}
    \caption{A situation where looking at each neuron in isolation is insufficient}
    \label{fig:beyond.z.value}
\end{figure}

Provided that an input $\vec{in}$ is drawn from a probability distribution, the value of $f^{1\rightarrow l}_i(\vec{in})$ also follows a distribution.
Consider the \emph{imaginary scenario} where $f^{1\rightarrow l}_i(\vec{in})$ has the Gaussian distribution with mean~$\mu$ and variance~$\sigma^2$, as demonstrated in Fig.~\ref{fig:standard.deviation}. Based on the interval estimate, around~$99.7\%$ of the values lie within three standard deviations of the mean. Therefore, one can compute the Z-score (Eq.~\ref{eq.z.score}) and use a simple containment checking if the Z-score falls within a specified interval such as $[-3, 3]$ for characterizing three standards of deviation. 

\begin{equation}\label{eq.z.score}
    z \defeq \frac{f^{1\rightarrow l}_i(\vec{in}) - \mu}{\sigma}
\end{equation}

Nevertheless, the method of using Z-values does not capture the interrelations among neurons. An example can be observed in Fig.~\ref{fig:beyond.z.value}, where for the value $f^{1\rightarrow l}(\vec{in_{ood}})$ is projected onto the plane with the axis on the $i_1$ and $i_2$ neuron values, correspondingly. If we only look at the Gaussian distribution in each dimension, then both $f^{1\rightarrow l}_1(\vec{in_{ood}})$ and $f^{1\rightarrow l}_2(\vec{in_{ood}})$ are considered to be in the decision boundary, as their values are within the minimum and maximum possible values when projecting the eclipse to the axis. However, it should be considered as OoD. The authors in~\cite{lee2018simple} thus considered \emph{Mahalanobis distance} (i.e., the distance is measured wrt.\ the scalar product induced by the \emph{inverse} of the positive-definite covariance matrix of the multi-variate Gaussian distribution, which transforms the covariance ellipsoid back into an unbiased sphere),
the generalization of the Z-score in the multi-variate Gaussian distribution setup, to characterize the distance measure for rejecting an input.

\subsection{DNN monitoring techniques from the FM community}

We now detail some of the monitoring techniques rooted in the formal methods community, where the key differences compared to the work from the ML community can be understood in one of the following dimensions.

\begin{itemize}
    \item Use \emph{abstraction} to decide the decision boundary, to ensure that all encountered data points within the data set are considered as in-distribution. 
    \item Move beyond convex decision boundaries and embrace \emph{non-convex decision} boundaries (via disjunction of sets), to allow more effective filtering of inputs being out-of-distribution. 
\end{itemize}

 \subsubsection{Abstraction-based monitoring using binary neuron activation patterns}
In 2018, Cheng, Nührenberg, and Yasuoka considered the monitoring problem for detecting abnormal inputs via activation pattern~\cite{cheng2019runtime}. The motivation is that  abnormal inputs shall enable DNN to create abnormal decisions, and one natural way of understanding the decision mechanisms is to view the activation of neurons in a layer as a binary word (feature activation), with each bit characterizing a particular neuron's binary \emph{on-off activation}. Conceptually, during run-time, when encountering an unseen feature activation pattern, one may reasonably doubt if the data collection has been insufficient or the input is OoD.

\vspace{2mm}
Precisely, for $x \in \mathbb{R}$, define the binary abstraction function  $b_{\alpha}(x)$ as follows. One natural selection of $\alpha$ value is considering $\alpha = 0$, which matches how a ReLU activation function performs suppression of its input. 

\begin{equation}
    b_{\alpha}(x) = 
\begin{cases}
  1 & \text{if $x> \alpha$} \\
  0 & \text{otherwise}
\end{cases}
\end{equation}

 Then given the vector $f^{1\rightarrow l}(\vec{in})$, the \emph{binary word vector} $bv_{\alpha}(f^{1\rightarrow l}(\vec{in}))$  
 is defined by element-wise application of $b_{\alpha}(\cdot)$ over $f^{1\rightarrow l}(\vec{in})$, i.e., 
\[
bv_{\alpha}(f^{1\rightarrow l}(\vec{in})) \defeq (b_{\alpha}(f^{1\rightarrow l}_1(\vec{in})), b_{\alpha}(f^{1\rightarrow l}_2(\vec{in})), \ldots, b_{\alpha}(f^{1\rightarrow l}_{d_l}(\vec{in})))
\]

Given $\mathcal{D}$, the monitor $ \mathcal{M}^{bv}$ is simply the set of all binary word vectors as detailed in Eq.~\ref{eq.binary.word.monitor}. The decision mechanism of the monitor is simple: If there is an input $\vec{in} \in \mathbb{R}^{d_0}$ such that $bv_{\alpha}(f^{1\rightarrow l}(\vec{in})) \not\in \mathcal{M}^{bv}$, then consider $\vec{in}$ as OoD.   
\begin{equation}\label{eq.binary.word.monitor}
    \mathcal{M}^{bv} \defeq \{ bv_{\alpha}(f^{1\rightarrow l}(\vec{in})) \; | \; (\vec{in}, \vec{label}) \in \mathcal{D}\}
\end{equation}

The question that immediately arises is how the containment checking (i.e., whether $bv_{\alpha}(f^{1\rightarrow l}(\vec{in})) \in \mathcal{M}^{bv}$ holds) can be made time-efficient. The authors in~\cite{cheng2019runtime} represent the set $\mathcal{M}^{bv}$ using \emph{binary decision diagrams} (BDDs)~\cite{bryant1992symbolic}. The containment checking can be done in time \emph{linear} to the number of monitored neurons. In order to reduce the size of the BDD and thus the memory footprint, various heuristics for variable re-ordering\footnote{BDDs are minimal acyclic finite automata that accept the fixed-length binary representation of some finite set wrt.\ a fixed order on the bits; finding an optimal order is NP-complete in general.}  can be applied before deploying the monitor. 

The second issue is to consider some slight variations in the created binary word, i.e., instead of rejecting every word not in $\mathcal{M}^{bv}$, one may relax the constraint only to reject if $bv_{\alpha}(f^{1\rightarrow l}(\vec{in}))$ has a Hamming distance greater than $\kappa$ to every word in $\mathcal{M}^{bv}$.
This can be easily realized in BDD by directly building the set of all words with Hamming distance to any word in $\mathcal{M}^{bv}$ being less or equal to $\kappa$~\cite{cheng2019runtime}.

Finally, one can also create more fine-grained decisions, using two binary variables to characterize four intervals rather than using one variable that can only split the domain into two, as detailed in an extension~\cite{cheng2021provably}. One limitation of using more variables for encoding one neuron is that the concept of Hamming distance can no longer be mapped with physical interpretations.

\subsubsection{Binary word monitoring without BDD, and abstraction on convolutional layers} The above basic principle on binary word encoding has been extended, 
where within the ML community, a joint team of academia and industry recently uses binary word monitoring to create the state-of-the-art OoD detectors against other techniques~\cite{olber2023CVPR}. The key innovation in~\cite{olber2023CVPR} is to utilize hardware accelerators such as GPU, where instead of building a BDD that compactly represents the set of binary words, simply store all binary words as 2D arrays/tensors (via libraries such as \texttt{np.ndarray} for numpy or \texttt{tf.tensor} for tensorflow). Then the containment checking is done by a hardware-assisted parallelized checking if one of the binary vectors in the tensor matches $bv_{\alpha}(f^{1\rightarrow l}(\vec{in}))$. Computing the Hamming distance can be implemented with these libraries by first applying an XOR operation, followed by counting the number of discrepancies. Another innovation is that the authors apply the binary word generation on convolutional layers, each of which may contain multiple channels. For example, we have one channel for a grayscale image and three channels for a color image, respectively. 
The work adopts adaptive pooling for every channel to select the most critical features and obtains a vector for all the channels in the convolutional layer. Finally, the vector is converted into a binary pattern with an adaptive activation function, using p-percentile values for the threshold.

 \subsubsection{Abstraction-based monitoring using the range of neuron values}

\begin{figure}[t]
    \centering
\includegraphics[width=\textwidth]{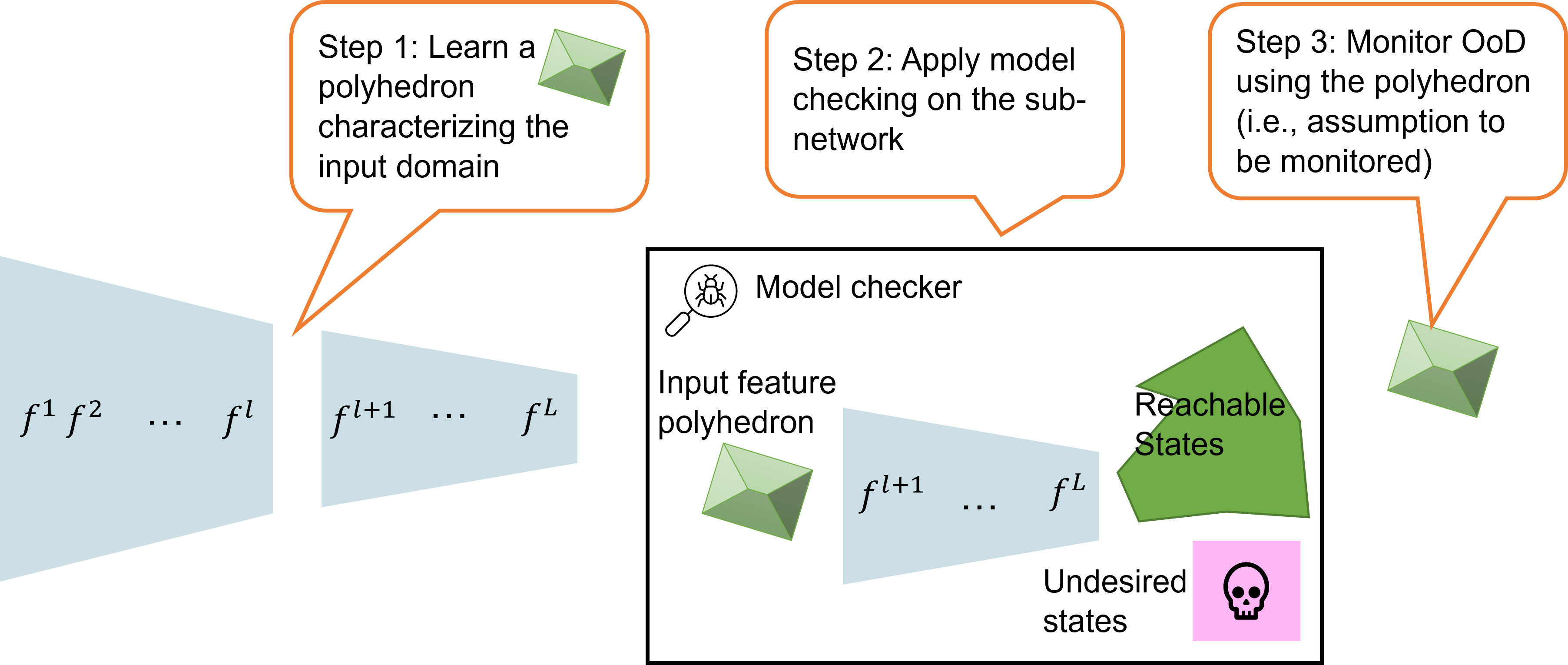}
    \caption{The role of box abstraction monitors in assume-guarantee-based verification of neural networks}
    \label{fig:assume.guarantee}
\end{figure}

Another idea of abstraction-based monitoring is to use the range of each neuron. This idea occurred independently in two papers around the same time~\cite{cheng2020towards,henzinger2019outside}, where we detail the underlying idea in~\cite{cheng2020towards}.  In~\cite{cheng2020towards}, the boxed abstraction monitor is introduced due to the need to perform \emph{assume-guarantee-based formal verification}. For formal verification of perception-based deep neural networks where the input dimension is extremely high (e.g., images or lidar point clouds), one encounters both scalability and specification challenges. The network can be too large to be verified. At the same time, we may not be interested in every input $\vec{in}\in \R^{d_0}$ but rather the set of inputs characterizing the human-specified operational domain. As illustrated in Fig.~\ref{fig:assume.guarantee}, the authors in~\cite{cheng2020towards} thus considered building an \emph{abstraction-based monitor} $\mathcal{M}^{box}$ using layer~$l$ that ensures to include all input $\vec{in}$ with $(\vec{in}, \vec{label}) \in \mathcal{D}$, the decision mechanism of~$\mathcal{M}^{box}$ shall not view $\vec{in}$ as OoD.

Precisely, the monitor $\mathcal{M}^{box} \defeq ([m_1, M_1] , \ldots, [m_{d_l}, M_{d_l}])$, where for $i \in \{1, \ldots, d_l\}$, $m_i$ and $M_i$ are defined using Eq.~\ref{eq.box.monitor}, with $\delta$ being a small positive constant that can be tuned. 

\begin{equation}\label{eq.box.monitor}
\begin{split}
 m_i  \defeq \min(\{f^{1\rightarrow l}_i(\vec{in}) \;|\;(\vec{in}, \vec{label}) \in \mathcal{D}\}) - \delta 
 \\
 M_i  \defeq  
 \max(\{f^{1\rightarrow l}_i(\vec{in}) \;|\; (\vec{in}, \vec{label})  \in \mathcal{D}\}) + \delta   
 \end{split}
\end{equation}

Then $\mathcal{M}^{box}$ considers $\vec{in}$ to be OoD if $\exists i$ such that $f^{1\rightarrow l}_i(\vec{in}) \not \in [m_i, M_i]$. In layman's words, the monitor is constructed by recording, for each neuron, the largest and smallest possible valuation (plus adding some buffers) that one can obtain when using the data set~$\mathcal{D}$.

The creation of the monitor $\mathcal{M}^{box}$ can be used in the assume-guarantee-based formal neural network verification as follows. Given a set of unsafe output states $\mathcal{S}_{risk} \subseteq \R^{L}$, the assume-guarantee-based formal verification first poses the following safety verification problem:

\begin{multline}
      \exists \vec{v} = (v_1, \ldots, v_{d_{l}}) \in \R^{d_l}  \;  
      \text{s.t.}   \; 
v_1 \in [m_1, M_1] \; \wedge \ldots \wedge v_l \in [m_l, M_l] \\
      \wedge 
f^L(f^{L-1}(\dots f^{l+1}(\vec{v}))) \in \mathcal{S}_{risk} 
\end{multline}

The verification problem is \emph{substantially simpler} than a standard DNN formal verification problem, as it only takes part of the DNN into analysis without considering high-dimensional inputs and layers~$f_1$ to~$f_l$. However, the safety guarantee of no input $\vec{in} \in \R^{d_0}$ generating an unsafe output is only conditional to the assumption where $\forall i: f^{1\rightarrow l}_i(\vec{in}) \in [m_i, M_i]$, which is monitored during runtime.

\paragraph{(Remark)} While one can observe that the boxed-abstraction is highly similar to the Z-value approach as described in Section~\ref{subsec.monitoring.ml}, the approach how decision boundary is created is conceptually different. The abstraction-based monitors \emph{do not assume any distribution} on the values of the neuron but demand a full enclosure for all neuron values from the data set.

\vspace{3mm}

\begin{figure}[t]
    \centering
\includegraphics[width=0.7\textwidth]{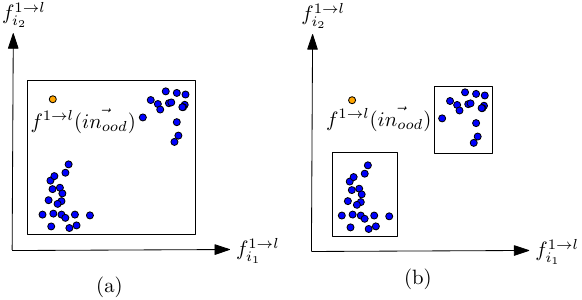}
    \caption{An example where using two boxes is more appropriate}
    \label{fig:beyond.single.box}
\end{figure}

\begin{figure}[htp]
		
		\begin{subfigure}{.95\textwidth}
  \centering
  \includegraphics[width=.9\linewidth]{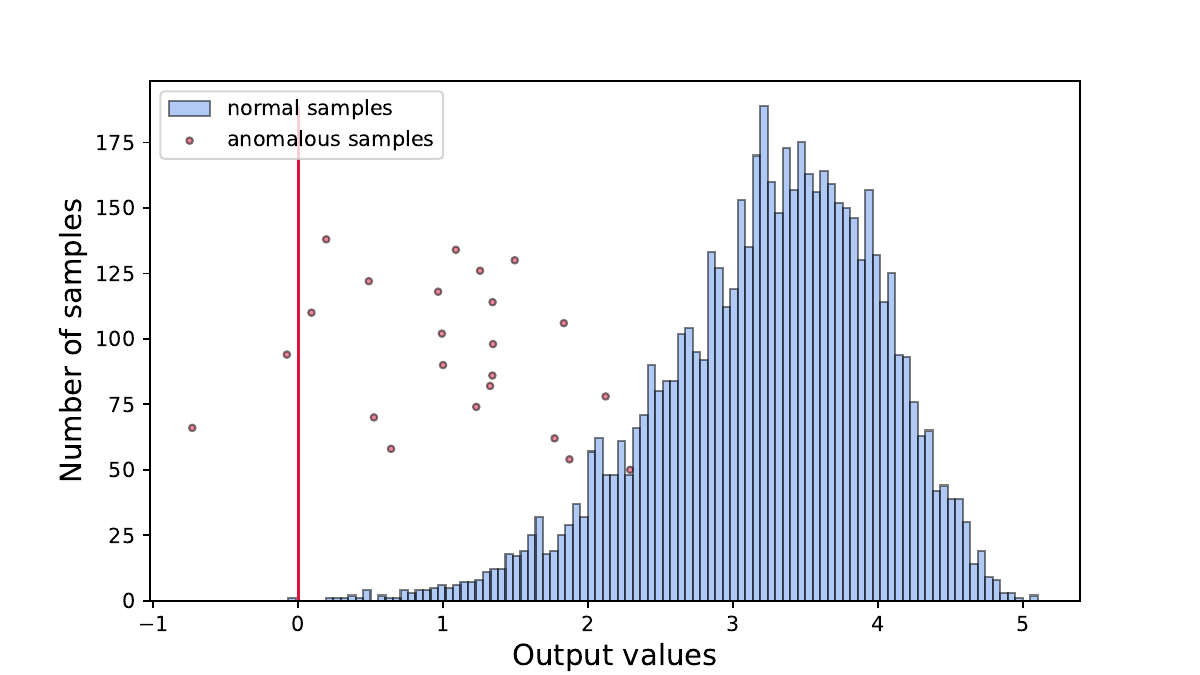}
  \caption{Values of an active neuron}
  \label{fig:emp1}
\end{subfigure}\\%
\begin{subfigure}{.95\textwidth}
  \centering
  \includegraphics[width=.9\linewidth]{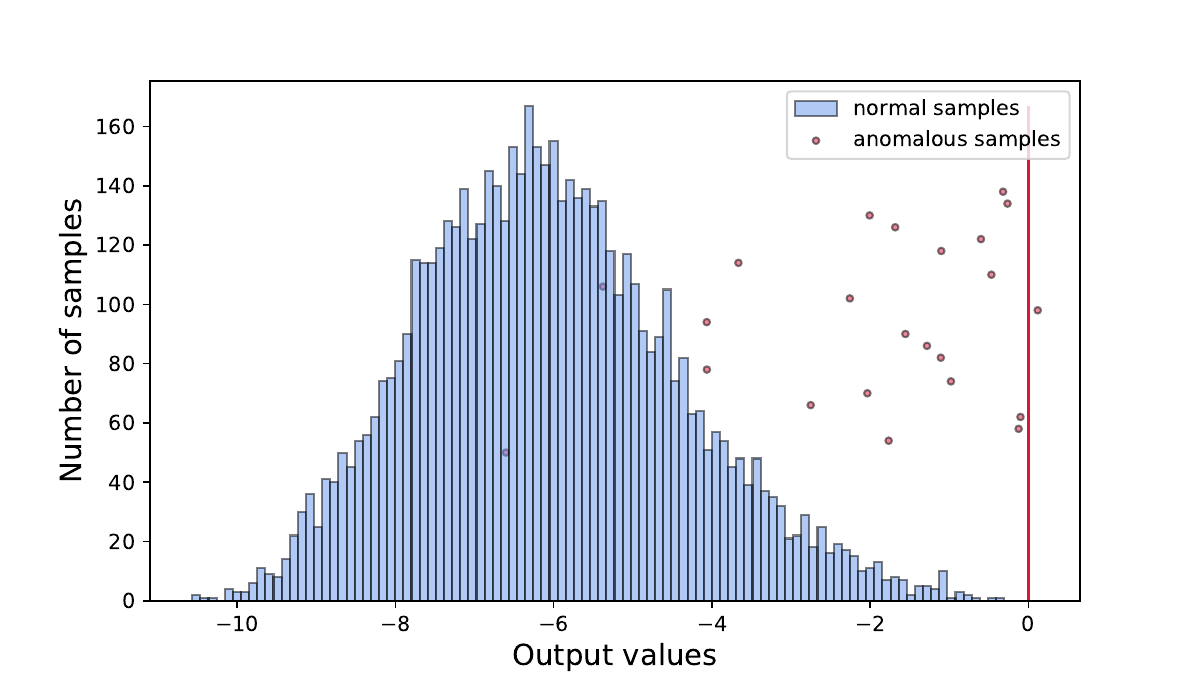}
  \caption{Values of an inactive neuron}
  \label{fig:emp0}
\end{subfigure}
\begin{subfigure}{.95\textwidth}
  \centering
  \includegraphics[width=.9\linewidth]{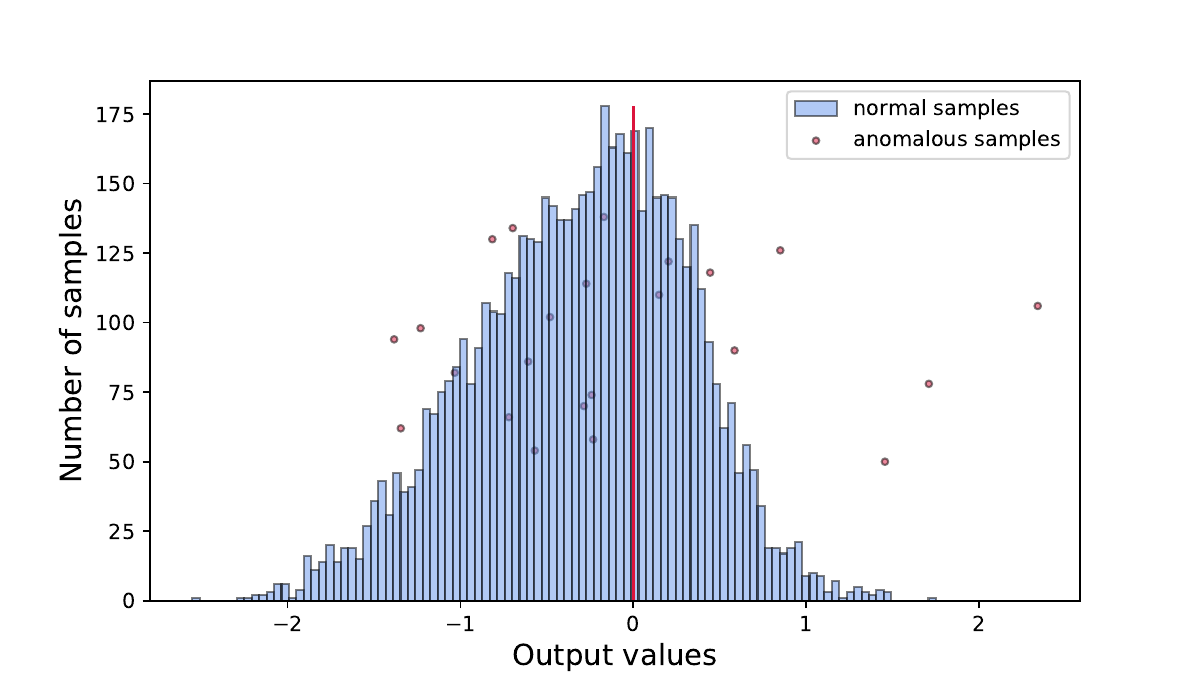}
  \caption{Values of an ignored neuron}
  \label{fig:emp2}
\end{subfigure}%
		\caption{Neuron value distribution before ReLU with data in the same class~\cite{yan2022priority}\label{fig:empirical}}
\end{figure}

\subsubsection{Extensions of Boxed Monitors} 

In the following, we highlight a few extensions that researchers in the formal method community created. 

Boxed abstraction also suffers from the precision problem.  Illustrated in Fig.~\ref{fig:beyond.single.box}(a), by only recording the minimum and maximum of each neuron value,  using only one box leads to including $f^{1\rightarrow l}(\vec{in_{ood}})$. The authors in~\cite{wu2021customizable} thus consider an extension where unsupervised learning is first applied to perform clustering, followed by building boxed abstraction monitors for every cluster, as illustrated in Fig.~\ref{fig:beyond.single.box}(b). The extension thus allows the set of points in the feature space to be considered as ``in-distribution" to be a non-convex set, similar to the capabilities of using binary neuron activation patterns.

Towards the concept of non-convex set, the work by Hashemi et al.~\cite{hashemi2021gaussian} also starts with a concept similar to that of the Z-value as specified in Section~\ref{subsec.monitoring.ml}. However, instead of directly rejecting an input $\vec{in}$ if $\exists i$ such that $f^{1\rightarrow l}_i(\vec{in}) \not \in [\mu_i - \kappa \sigma_i, \mu_i + \kappa \sigma_i]$, the method only reports OoD when the number of violations is larger than a constant $\alpha$. For instance, $\alpha = 2$ implies that the monitor can tolerate the number of neuron range violations by up to two neurons. 

In the DNN-based classifiers, the values of some neurons with a class may always be greater than zero, while the value distribution of other neurons may be quite different.  Inspired by such observation, zero is usually regarded as the threshold to decide whether a neuron contributes to the pattern of a given class~\cite{zhangzzgy020}. 

Finally, the methods presented are largely generic without detailing what layer is appropriate to monitor and treatments on neuron activation patterns on in-distribution but wrongly predicted inputs. Another extension is to consider the distribution of neuron values before its activation function to select the candidate neurons for pattern representation~\cite{yan2022priority}. Fig.~\ref{fig:empirical} provides three types of neuron distributions for the samples in the same class. The blue column shows the number of samples leading to the corresponding values, and the red dot shows the values from wrong predictions. The method selects the neurons whose values are always larger than or less than zero for pattern encoding. The pattern of a class consists of neurons and the upper or lower bound to avoid the influence of values from wrong predictions in the given training set. Given an input during run-time, it counts the number of neurons deviating from the specified bounds. If the number is larger than a given threshold, the input is regarded as abnormal.

\subsection{Monitoring techniques without analyzing the DNN}
Except for extracting the patterns of DNNs or the distribution of output confidence, some works consider ML-based components as a black box and monitor abnormal outputs. 

\subsubsection{Consistency-based monitoring} 
Some works monitor the  consistency between detected objects from image streams. For example, PerceMon~\cite{balakrishnanDHY21} adopts specifications in timed quality temporal logic and its extensions with spatial operators to specify the properties. It monitors whether the data stream extracted from the ML-based component satisfies the specified property. The data stream includes the positions, classification, and detection confidence of objects. The structure of the monitor is similar to an existing online monitoring tool RTAMT~\cite{nivckovic2020rtamt} for STL specifications. The performance of the monitor decreases rapidly with the increasing number of objects in a frame.   

Another work considers the consistency in image streams and the consistency with the training data sets~\cite{chen2021runtime}. Given the training data set and the associated labels, it divides an image into various regions, collects the attributes of objects for every region, and constructs the dictionary. During runtime,  if the location or the size of an object detected in a frame is not in the dictionary, the alarm is triggered. The consistency between various image frames (temporal relation) helps to locate abnormalities such as label flips and object loss.

\subsubsection{Learning and monitoring operational design domains (ODD)}

Using the training set to estimate the expected distribution of the inputs and/or the outputs is one form of defining (some aspect of) the \emph{operational design domain} the ML-based component is supposed to operate in.
A formal description (i.e., a specification in some logic) $\phi'$ of the complete (human-defined) ODD is often not available or infeasible, or only implicitly given by means of the training set (e.g.\ pre-classified by a human).
That is, in general, we only have an incomplete specification $\phi$ of the ODD, which is an over-approximation (or relaxation) with $\models(\phi'\to \phi)$ (as a special case, we have $\phi=\textsf{true}$); further $\phi$ might refer to properties (events) which are not observable in the actual implementation. Different variants of refining such an initial incomplete specification $\phi'$ using the training set or the ML-based component have been proposed recently:

For instance, in~\cite{DBLP:conf/atva/TorfahXJVS22} the authors treat the already trained ML-based system (e.g.\ a controller) as a black box containing the ODD specification.
By interacting with the black box in an simulated environment (using \url{https://github.com/BerkeleyLearnVerify/VerifAI}), a monitor (in case of~\cite{DBLP:conf/atva/TorfahXJVS22} a decision tree) for detecting OoD-behavior is learned. Conformance testing is used to discover counterexamples which trigger additional simulations. 
A specification in MTL \emph{for the simulation} (which can thus refer to events that are not observable to the black box and/or which are supposed to be inferred by it) allows to guide the generation of the traces used for training and checking the monitor.

The authors in~\cite{DBLP:conf/iccps/LindemannQDP23} propose to train e.g.\ an LSTM (long short-term memory) to predict the future extension of the current run (trajectory) up to some finite horizon, and derive bounds (based on the training and calibration set) on the \emph{quantitative semantics} to decide whether a given specification $\phi$ will be satisfied by the future run at least with probability $1-\delta$ as long as the prediction is within the computed bounds. The specification is assumed to be given in bounded STL whose quantitative semantics returns a real number measuring how ``robust'' the satisfaction is wrt.\ small changes of the prediction. The requirement that the specification has to be bounded translates into the existence of a finite horizon up to which it suffices to predict future behavior.

Somewhat related is the idea to use the human-specified ODD already during the actual training. E.g.\ the authors in~\cite{DBLP:conf/atva/WagaCPKTH22} study how to avoid in reinforcement learning (RL) that the agent is trained on unsafe or unwanted behavior: here, one usually unknown part of the ODD is the environment with which the agent is supposed to interact.
The authors therefore use a specification of central safety aspects to not only \emph{shield} the agent from being trained on unsafe behavior, but also to learn an approximate model of the environment to guide the exploration of the environment during RL and further improve the shield.
The approximate model itself is obtained by combining learning algorithms for formal languages and computing optimal strategies for two-player games, similar to the approach used in formal synthesis. The computed strategy translates into a Mealy machine, which is then used for shielding the RL.

\subsubsection{Entropy-based monitoring} As the probabilities provided by the softmax function may be overconfident to out-of-distribution data, some works analyze the distribution of the probabilities from the softmax layer to separate out-of-distribution data from in-distribution data.  Entropy-based techniques aim to measure the uncertainty of the probabilities for all the classes. The larger entropy indicates higher uncertainty. For example, one can compute Shannon entropy (Eq.~\ref{eq.shannon}) to decide whether the prediction is acceptable, where \textbf{p} is the predictive distribution of the softmax layer. 
    \begin{equation}\label{eq.shannon}
        H(\textbf{p})=-\sum_{j}p_j \log p_j
    \end{equation}

In addition to Shannon entropy, the generalized entropy can also be applied to measure the uncertainty of the prediction (Eq.~\ref{eq.generalized})~\cite{liu2023gen}.
    \begin{equation}\label{eq.generalized}
        G_{\gamma}(\textbf{p})=\sum_{j}p^{\gamma}_j(1-p_j)^{\gamma}
    \end{equation}
 where      $\gamma\in (0,1)$. Lower values of $\gamma$ are more sensitive to uncertainties in the predictive distribution. 
 The intuition of the generalized entropy is to amplify minor derivations of a predictive distribution from the ideal one-hot encoding. 

    Facing applications with many classifications, the large fraction of very small predictive probabilities may have a significant impact on the generalized entropy. In such a case,  one can consider the top-M classes to capture small entropy variations in the top-M classes. 

    Finally, the score from entropy can be combined with statistics from training data to improve the performance of out-of-distribution data detection. However, the performance from the combination of various techniques may not always be the best due to the variety of data and model architecture.

\section{Challenges Ahead}

There is no doubt that monitors are instrumental in realizing safety-critical perception systems. This tutorial outlined the special challenges in monitoring perception systems, followed by a sketch of some notable monitoring techniques proposed by the machine learning community and developments from the formal methods community.  

Despite many fruitful results with ongoing technical innovations for DNN-based perception monitoring, we observe the absence of a rigorous design approach for developing, verifying, and validating OoD detectors. Such design principles must be carefully tailored to match the intended functionality and the specific operational domain they are meant to serve. This includes dimensions such as principled data collection (a counterexample is the evaluation conducted in this tutorial; we use Fashion-MNIST as the only OoD examples for MNIST),  using reasonable assumptions (e.g., instead of arbitrarily assuming the neuron values have a Gaussian distribution and apply the $3\sigma$ rule for characterizing $99.7\%$ of the data, use the more conservative Chebyshev's inequality that is applicable for arbitrary distribution), or the rigorous design of decision boundaries (e.g., consider the joint distribution of the output neurons, use better statistical methods for estimating their joint distribution; use higher moments than expected value and variance).

\bibliographystyle{abbrv}

\end{document}